\documentclass[letterpaper, 10 pt, conference]{ieeeconf}  

\IEEEoverridecommandlockouts                              

\binoppenalty=\maxdimen
\relpenalty=\maxdimen

\usepackage{graphicx}
\usepackage{bm}
\usepackage{amsmath}
\usepackage{amssymb}
\usepackage{amsfonts}
\usepackage{esvect}
\usepackage{textcomp}
\usepackage{booktabs}
\usepackage{tabularx}
\usepackage{float}
\usepackage{xcolor}
\usepackage{siunitx}
\usepackage{lipsum}
\usepackage{subfig}

\newcommand*\squeezespaces[1]{
  \thickmuskip=\scalemuskip{\thickmuskip}{#1}%
  \medmuskip=\scalemuskip{\medmuskip}{#1}%
  \thinmuskip=\scalemuskip{\thinmuskip}{#1}%
  \nulldelimiterspace=#1\nulldelimiterspace
  \scriptspace=#1\scriptspace
}
\newcommand*\scalemuskip[2]{%
  \muexpr #1*\numexpr\dimexpr#2pt\relax\relax/65536\relax
} 

\usepackage{sidecap}
\graphicspath{{./figures/}}

\newcommand{\xx}{\bm{\mathrm{x}}} 
\newcommand{\ww}{\bm{\mathrm{w}}} 
\newcommand{\yy}{\bm{\mathrm{m}}^*} 
\newcommand{\oo}{\bm{\mathrm{y}}}  
\newcommand{\zx}{\bm{\mathrm{\epsilon}}}
\newcommand{\zw}{\bm{\mathrm{\kappa}}}
\newcommand{\px}{p_{\pfx}}
\newcommand{\pw}{p_{\pfw}}

\newcommand{\fx}{g_{\pfx}}
\newcommand{\fw}{g_{\pfw}}
\newcommand{\fo}{y_{\pfo}}
\newcommand{\dx}{d_{\pdx}}
\newcommand{\dw}{d_{\pdw}}
\newcommand{\pzx}{p_{\mathrm{\epsilon}}}

\newcommand{\pfx}{\theta_{\mathrm{x}}}
\newcommand{\pfw}{\theta_{\mathrm{w}}}
\newcommand{\pfo}{\alpha}
\newcommand{\pdx}{\beta_{\mathrm{x}}}
\newcommand{\pdw}{\beta_{\mathrm{w}}}
\newcommand{\Dr}{\mathtt{R}}
\newcommand{\Ds}{\mathtt{S}}

\newcommand{\tick}{\checkmark}

\renewcommand{\baselinestretch}{0.945}

\title{\LARGE \bf
There and Back Again: Learning to Simulate Radar Data for Real-World Applications}

\author{Rob Weston, Oiwi Parker Jones and Ingmar Posner$^{*}$ 
\thanks{$^{*}$Applied Artificial Intelligence Lab (A2I), University of Oxford
        {\tt\small \{robw, oiwi, ingmar\}@robots.ox.ac.uk}%
}
}

\begin{document}

\maketitle
\thispagestyle{empty}
\pagestyle{empty}

\begin{abstract}
Simulating realistic radar data has the potential to significantly accelerate the development of data-driven approaches to radar processing. However, it is fraught with difficulty due to the notoriously complex image formation process. Here we propose to learn a radar sensor model capable of synthesising faithful radar observations based on simulated elevation maps. In particular, we adopt an adversarial approach to learning a \emph{forward} sensor model from unaligned radar examples. In addition, modelling the \emph{backward} model encourages the output to remain aligned to the world state through a cyclical consistency criterion. The backward model is further constrained to predict elevation maps from real radar data that are grounded by partial measurements obtained from corresponding lidar scans. Both models are trained in a joint optimisation. We demonstrate the efficacy of our approach by evaluating a down-stream segmentation model trained purely on simulated data in a real-world deployment. This achieves performance within four percentage points of the same model trained entirely on real data. 

\end{abstract}
\section{Introduction}
\label{sec:intro}
The long sensing range of radar and its resilience to adverse environmental conditions make it an attractive complement to lidar and vision for robotics and autonomous driving applications. However, radar is notoriously challenging to interpret; multi-path phenomena, limited resolution, and pernicious noise artefacts arising throughout the complex and imperfect measurement pipeline pose significant challenges to radar-based perception systems. In recent years, data-driven approaches have made significant strides to overcome these challenges across a range of tasks in robotics \cite{weston2019probably, barnes2019masking, KidnappedRadarArXiv, tang2020rsl}. Central to the continuing success of such approaches is the quality, scale, and labelling of radar datasets, which in contrast to vision and lidar remain limited.

Akin to progress in the use of other sensing modalities, simulation has the potential to significantly accelerate the development and deployment of radar-based methods by reducing the need for human annotation and automating the data-gathering process. The importance of learning from simulated data can be seen across a wide range of tasks in vision and lidar \cite{manivasagam2020lidarsim, sun2018pwc, mayer2016large, ros2016synthia, richter2016playing, gaidon2016virtual}; and it is echoed in the rapid development of multiple autonomous driving simulators capable of simulating complex worlds, designed to facilitate these approaches \cite{dosovitskiy2017carla, wymann2000torcs, airsim2017fsr}.

\begin{figure}[ht]
    \centering
    \includegraphics[width=\linewidth]{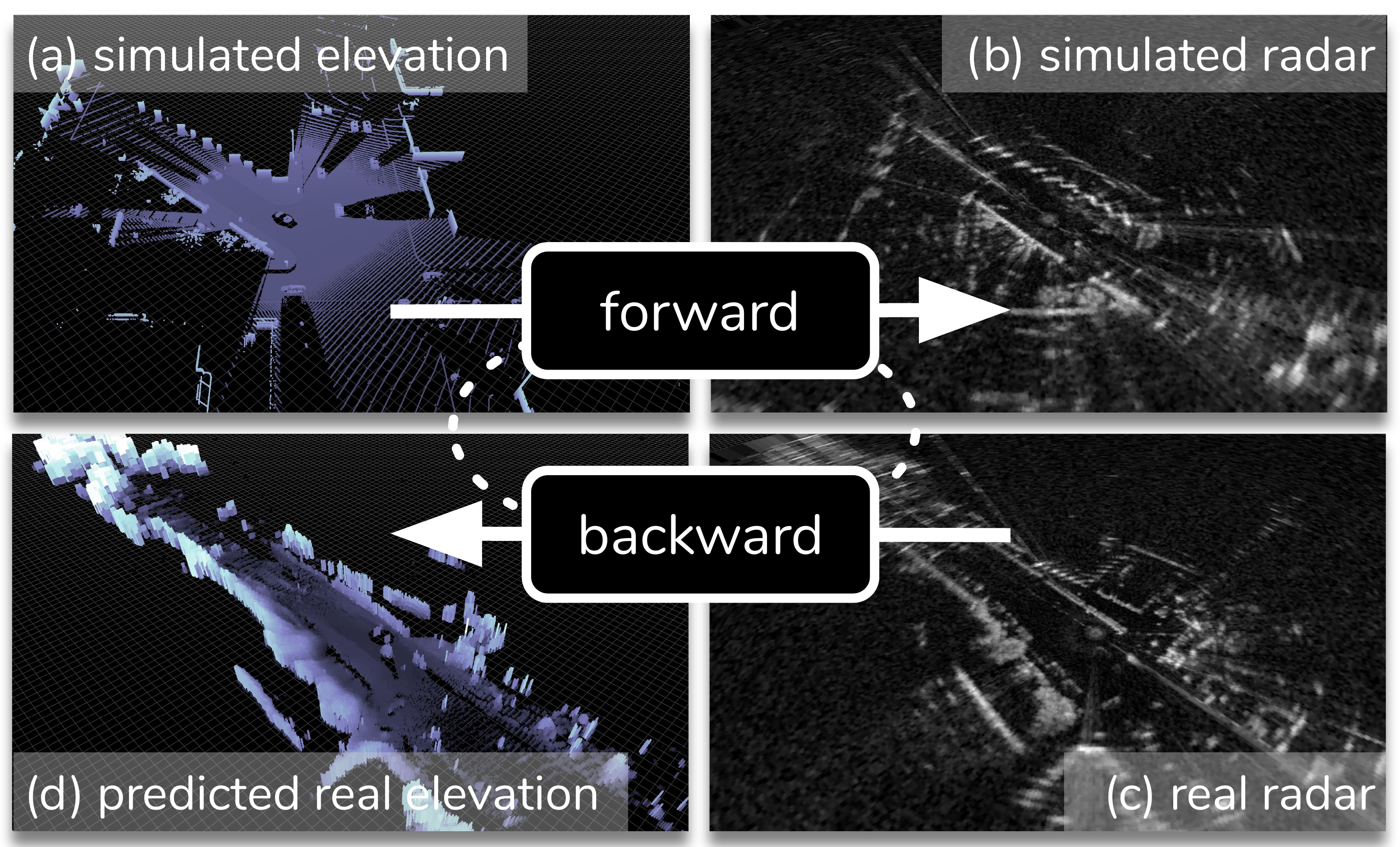}
    \caption{Given a simulated elevation map (a) we are able to generate realistic radar in simulation (b) through a data driven approach. We achieve this by learning from unaligned real radar observations (c). Alongside the forward mapping we also learn the backward mapping from real radar (c) to predict the real world elevation (d). This allows us to further constrain training through cyclical consistency and by learning from partial lidar measurements collected in the real world.}
    \label{fig:teaser}
    \vspace{-3em}
\end{figure}

Inspired by the impact simulation has brought to the development of sophisticated vision and lidar systems, the overarching goal of our work is to enable the training of data-driven models for radar interpretation in simulation. To successfully train models in simulation, we therefore consider learning a radar sensor model that is able to faithfully simulate radar observations, such that the domain gap experienced by a down-stream system when trained on either real or synthetic data is minimal. In particular, we aim to interface with existing simulators already capable of synthesising complex scenes and adopted widely by the community. In meeting this requirement, we consider simulating radar observations given a layout of the world, supplied to our model in the form of an elevation map. Noise processes arising throughout the radar sensing pipeline give radar an inherently stochastic nature; to accurately replicate this, we adopt a probabilistic approach, using a \emph{deep implicit model} \cite{mohamed2016learning} to capture a distribution over possible radar sensor measurements.

Adopting an adversarial paradigm, we train our sensor model to explicitly generate realistic radar observations from \emph{simulated} height maps using unaligned real-world radar data. As shown in Figure \ref{fig:teaser}, alongside the forward sensor model we also learn the backward model to infer the elevation state of the real world from radar. This allows us to further constrain training by enforcing cyclical consistency \cite{CycleGAN2017} between the forward and backward processes. Alongside this we ground the predictions of our backward model in the real world by enforcing alignment with partial height measurements generated automatically from lidar.

Through our approach, we demonstrate the feasibility of training radar models in simulation to the best of our knowledge for the first time. In doing so, our approach achieves performance within $4\%$ compared to that of the same model trained in the real world.
\section{Related Work}
Whilst unexplored in radar, the benefits of training in simulation have been extensively studied in both vision \cite{gaidon2016virtual, richter2016playing, ros2016synthia, mayer2016large, sun2018pwc, kim2017learning} and lidar \cite{manivasagam2020lidarsim} across a range of perception tasks including segmentation \cite{gaidon2016virtual, ros2016synthia}, detection \cite{manivasagam2020lidarsim}, tracking \cite{gaidon2016virtual}, and optical flow \cite{sun2018pwc}. In addition to training models, simulation has an important role to play in the analysis and interrogation of corner-cases, potentially too dangerous to recreate in the real world. The need for simulation, in combination with the development of powerful open source games engines \cite{unrealengine, haas2014history}, has led to a profusion of both open source \cite{wymann2000torcs, airsim2017fsr, dosovitskiy2017carla} and proprietary \cite{unrealengine} autonomous driving simulators for vision and lidar but support for radar still remains limited. Whilst \cite{dosovitskiy2017carla} provides a simple simulation of target lists as might be returned by a typical automotive radar, in this work we are concerned with simulating significantly richer---but more challenging to replicate---raw radar power measurements.

To capitalise on the similar successes that training in simulation has brought to vision and lidar, we aim to learn a radar sensor model capable of faithfully simulating the radar sensing process. But radar's complex sensing pipeline makes simulation challenging. The high frequency of radar renders direct solution of Maxwell's equations intractable. Instead, asymptotic solutions have been widely adopted, in an attempt to simulate radar, relying on a combination of geometric \cite{ling1989shooting} and physical optics \cite{weinmann2006ray, domingo1995computation, rius1993greco}. Representing objects as their characteristic scattering centers \cite{hurst1987scattering, schuler2008extraction, andres20133d} can further help reduce computation. Whilst several approaches have demonstrated the feasibility of simulating simple driving scenes using these methods \cite{chipengo2018antenna, chipengo201977, hirsenkorn2017ray, machida2019rapid}, they scale poorly, relying on a precise model of the world, including material properties that are difficult to model in practice. In contrast, our approach is capable of simulating radar observations given only a simulated elevation map; this allows us to interface with and scale to the complex worlds provided in modern simulation environments, without requiring us to build a radar-specific simulation world from the ground up.

Several approaches also propose to account for noise artefacts arising through the downstream measurement process using hand-crafted phenomenological models \cite{buhren2006simulation, buhren2007extension, buhren2007simulation}. More recently, data-driven approaches \cite{wheeler2017deep} have been proposed as a better alternative \cite{holder2018measurements}, learning to characterise the entire radar process, from world state to sensor observation, from raw data. Whilst \cite{wheeler2017deep} shows the feasibility of using a data-driven approach to replicate radar data in a controlled airfield environment with only a handful of targets, in our work we consider simulating radar in complex urban environments, where labelling exact real-world layouts is significantly more challenging. Instead of requiring exact real world layouts to train our model, as in \cite{wheeler2017deep}, we generate radar observations from simulated height maps. As well as side-stepping the need for precise layouts of the real world, through this approach we are able to explicitly encourage our model to generate feasible radar observations \emph{in simulation}. 

Similar in flavour, therefore, to recent approaches proposed for unaligned domain transfer in vision \cite{CycleGAN2017, kim2017learning, yi2017dualgan, yang2020surfelgan}, we too consider learning \emph{unaligned} mappings between simulated world layouts and real radar observations. We model the forward and backward model side-by-side using adversarial and cyclical consistency losses. Unlike in \cite{CycleGAN2017, kim2017learning, yi2017dualgan, yang2020surfelgan} which consider a deterministic one-to-one mapping between domains, we adopt an inherently probabilistic approach. We capture a \emph{distribution} over possible power returns to account for the stochastic noise processes arising throughout the radar sensing pipeline. We also further encourage our backward model to predict elevation states that are aligned to real-world measurements generated automatically in lidar, where they are available. 
\section{Deep Radar Simulation}
\label{sec:approach}

\subsection{Problem Formulation}
\label{sec:formulation}
Let $\xx^*$ denote a real radar observation generated from real world state $\ww^* \sim p(\ww^*)$ as
\begin{gather}
    \xx^* \sim p(\xx^*| \ww^*), \quad \ww^* \sim p(\ww^*)
\end{gather}
where $p(\xx^*| \ww^*)$ is the real radar sensor model. Specifically, we consider radar observations of the form $\xx^* \in \mathbb{R}^{\Phi \times R}$ where $\xx^*_{i j}$ gives the power returned from the world at polar-coordinate $(\mathrm{\phi}_i, \mathrm{r}_j) \in \{\phi_i\}_{i=1}^\Phi \times \{\mathrm{r}_j \}_{j=1}^R $.

We aim to learn a radar sensor model $\xx \sim \px(\xx | \ww)$ capable of generating feasible radar observations $\xx$ from a \emph{simulated} state $\ww \sim p(\ww)$. We model the world state $\ww \in \mathbb{R}^{\mathrm{\Phi} \times \mathrm{R}}$, with the same dimensions as $\xx$, where $\ww_{i, j}$ gives the elevation of the world location $(\phi_i, r_i)$. In addition to containing the necessary information to generate $\xx$ (such as the shape, size, and position of objects in the scene), this representation allows us to easily interface with a preexisting simulation environment, such as CARLA \cite{dosovitskiy2017carla}. Whilst $\ww$ is easily simulated it is more challenging to measure in the real world. We assume that we are able to obtain \emph{partial} measurements $\yy = m(\ww^*)$ using lidar, where due to the limited range and number of beams of lidar many locations $(\phi_i, r_i)$ are without an elevation measurement. We therefore have at our disposal real world observations $\Dr =\{(\xx^*, \yy)_n \}_{n=1}^N$ and simulated observations $\Ds = \{\ww_l \}_{l=1}^L$ with which to train our approach.

\subsection{Stochastic Simulation Using Deep Implicit Models}
\label{sec:approach-stochastic}
For the same world state $\ww^*$, aleatoric sensor noise results in radar sensor measurements $\xx^*_t \sim p(\xx^* | \ww^*)$ and $\xx^*_{t +1} \sim p(\xx^* | \ww^*)$ that are inherently stochastic. To capture the stochasticity in the mapping from $\ww$ to $\xx$ -- mimicking the true sensing process -- we introduce a latent variable $\zx \sim \mathcal{N}(\bm{0}, \bm{I})$; using a neural network $\fx(\ww, \zx)$ with parameters $\pfx$, for a fixed $\ww$, we are able to sample multiple possible $\xx$ by sampling $\zx \sim \mathcal{N}(\bm{0}, \bm{I})$. This allows us to implicitly capture a distribution $\px(\xx | \ww)$ over possible radar observations sampling $\xx \sim \px(\xx | \ww)$ as
\begin{equation}
    \label{eq:implicit}
    \xx = \fx(\ww, \zx; \pfx) \quad \text{with} \quad \zx \sim \pzx(\zx) \text{ .}
\end{equation}
Through this approach we are able to leverage the representational power of a deep neural network $\fx$ to learn the complex mapping from world state $\ww$ to radar sensor observation $\xx$, whilst simultaneously implicitly learning how to characterise the uncertainty in the sensing process.

\subsection{Learning only from Real Observations}
\label{sec:approach-real_only}
If we could observe real-world observation-state pairs $\Dr=\{(\xx^*, \ww^*)_n  \}_{n=1}^N$ we might train $\px(\xx | \ww)$ by minimising
\begin{equation}
\mbox{$\squeezespaces{0.8} 
     \mathcal{A}_\mathrm{x} = \mathbb{E}_{p(\xx^*, \ww^*)}[\ell(\pfx; \xx^*, \ww^*)] \approx \frac{1}{N} \sum_{n} \ell(\pfx; \xx^*_n, \ww^*_n)
          $} \\
    \label{eq:align-loss}
\end{equation}
where $\ell(\pfx; \xx^*, \ww^*)$ is a regression loss (eg. MAE or MSE) between the real $\xx^*$ and simulated observations $\xx \sim \px(\xx | \ww^*)$. However, only partial partial measurements of the world state $\yy = m(\ww^*)$ are available. In addition, by training our model in this way, $\px(\xx | \ww)$ is only trained on real state observations $\ww^* \in \Dr$ and a significant domain gap between $\ww \in \Ds$ and $\ww^* \in \Dr$ is still likely to persist. We posit that models that explicitly incorporate simulated state observations $\ww \in \Ds$ into the training loop are more likely to generate feasible radar observations in simulation. We now propose how this might be achieved.

\subsection{Learning from Unaligned Real and Simulated Data}
\label{sec:approach-adversarial}
We assume that measurements of the world state $\yy$ are entirely unavailable. In this case, we have the datasets $\Dr = \{\xx^*_n\}_{n=1}^N$ and $\Ds = \{\ww_n\}_{n=1}^N$ available to us. As $(\xx^*_n, \ww^*_n) \in \Dr \times \Ds$ are no longer aligned, training $\px(\xx| \ww)$ as in (\ref{eq:align-loss}) is no longer feasible.

To train $\xx \sim \px(\xx | \ww)$ to replicate $\xx^* \sim p(\xx^* | \ww^*)$ using only unaligned examples $(\xx^*, \ww)$ we adopt an adversarial approach \cite{goodfellow2014generative}; introducing a discriminator network $\dx(\xx)$ with parameters $\pdx$, and training objectives, 
\begin{gather}
    \label{eq:disc}
    \mathcal{D}_\mathrm{x} = \mathbb{E}_{p(\xx^*)}\big[ \big(\dx (\xx^*) - 1 \big)^2 \big] + \mathbb{E}_{\px(\xx)}\big[\dx(\xx)^2\big] \\
    \label{eq:gen}
    \mathcal{G}_\mathrm{x}  = \mathbb{E}_{\px(\xx)} \big[ \big(\dx(\xx) - 1 \big)^2 \big]
\end{gather}
minimising $\mathcal{D}_\mathrm{x}(\pdx)$ with respect to $\pdx$ and $\mathcal{G}_\mathrm{x}(\pfx)$ with respect to $\pfx$.\footnote{More generally the expectations in (\ref{eq:disc}) and (\ref{eq:gen}) could be written with respect to the joint $p(\xx, \ww)$ but as the terms inside the expectations are only functions of $f(\xx)$ we have $\mathbb{E}_{p(\xx, \ww)}[f(\xx)] = \mathbb{E}_{p(\xx)}[f(\xx)]$ which we adopt for simplicity. Here we consider generating samples $\xx \sim p(\xx)$ by sampling from the joint distribution $\xx, \ww \sim p(\xx, \ww)$ and disregarding $\ww$} Here we have adopted a least-squares loss assuming a $[\xx, \xx^*] = [0,1]$ coding scheme: this avoids discriminator saturation which can destabilise training \cite{mao2017}. 

In reality the expectations in (\ref{eq:disc}) and  (\ref{eq:gen}) are estimated using $\mathbb{E}_{p(\xx)}[f(\xx)] \approx \frac{1}{K} \sum_{k} f(\xx_k)$ with $\xx_k \sim p(\xx)$; crucially this allows us to train $\px(\xx | \ww)$ using only unaligned samples $(\xx^*, \ww) \in \Dr \times \Ds$ sampling from $\px(\xx)$ as $\xx_k \sim \px(\xx| \ww_k)$ using (\ref{eq:implicit}).

\subsection{Constraining Training Using Cyclical Consistency}
\label{sec:approach-cyc}
However, training $\px(\xx | \ww)$ by minimising just $\mathcal{G}_{\mathrm{x}}(\pfx)$ with unaligned examples is a highly unconstrained process \cite{CycleGAN2017}. In the worst case, $\px(\xx | \ww)$ could choose to disregard the state information $\ww$ entirely, instead generating $\xx = \fx(\ww, \zx)$ using only $\zx$ as in a standard GAN formulation \cite{goodfellow2014generative}. 

To counter this, in order to further constrain $\px(\xx | \ww)$, we also model the backward mapping $\ww^* \sim p(\ww^* | \xx^*)$ as $\ww = \fw(\xx, \zw; \pfw)$ with $\zw \sim \mathcal{N}(\bm{0}, \bm{I})$, where $\fw$ is a neural network with parameters $\pfw$ and $\zw$ induces the uncertainty in $p(\ww^*| \xx^*)$. Crucially, this allows us to impose additional cyclical consistency constraints \cite{CycleGAN2017, kim2017learning, yi2017dualgan},
\begin{align}
    \label{eq:consitency_w}
    \ww'' &\approx \ww & \ww'' &\sim \pw(\ww^* | \xx') & \xx' &\sim  \px(\xx | \ww) \\
    \label{eq:consitency_x}
    \xx'' &\approx \xx^* & \xx'' &\sim \px(\xx | \ww') & \ww' &\sim  \pw(\ww^* | \xx^*)
\end{align}
into our training framework, explicitly encouraging $\px(\xx | \ww)$ to use $\ww$ by enforcing (\ref{eq:consitency_w}) and $\pw(\ww | \xx)$ to use $\xx$ by enforcing (\ref{eq:consitency_x}) through cyclical consistency losses,
\begin{gather}
    \label{eq:cyc_w}
    \mathcal{C}_{\mathrm{w}}(\pfx, \pfw) = \mathbb{E}_{p(\ww)}\left[ \| \ww - \ww'' \|_{1}\right] \\
    \label{eq:cyc_x}
    \mathcal{C}_{\mathrm{x}}(\pfx, \pfw) = \mathbb{E}_{p(\xx^*)}\left[ \| \xx^* - \xx'' \|_{1} \right] \text{ .}
\end{gather}
Alongside (\ref{eq:cyc_w}) and (\ref{eq:cyc_x}), we also train $\pw(\ww^*| \xx^*)$ using an adversarial objective. Introducing another discriminator network $\dw(\ww)$ with parameters $\pdw$, and training objectives,
\begin{gather}
    \label{eq:disc_w}
    \mbox{$\squeezespaces{0.5} 
          \mathcal{D}_\mathrm{w} = \mathbb{E}_{p(\ww)}\big[ \big(\dw (\ww) - 1 \big)^2 \big] + \mathbb{E}_{\pw(\ww^*)}\big[\dw(\ww^*)^2\big]
          $} \\
    \label{eq:gen_w}
    \mathcal{G}_\mathrm{w}  = \mathbb{E}_{\pw(\ww^*)} \big[ \big(\dw(\ww^*) - 1 \big)^2 \big]
\end{gather}
we minimise $\mathcal{D}_\mathrm{w}(\pdw)$ with respect to $\pdw$ and $\mathcal{G}_\mathrm{w}(\pfw)$ with respect to $\pfw$, generating samples $\ww^*_k \sim \pw(\ww^*)$ as $\ww^*_k \sim \pw(\ww^* | \xx^*_k)$ using (\ref{eq:disc_w}) with $\xx^*_k \in \Dr \sim p(\xx^*)$.

We note that just as $\mathcal{G}_\mathrm{x}(\pfx)$ could lead $\fx(\ww, \zx)$ to ignore $\ww$ (as discussed previously), in the worst case $\mathcal{C}_\mathrm{x}(\pfx, \pfw)$ encourages $\xx =\fx(\ww, \zx)$ to ignore $\zx$ enforcing a one-to-one mapping between $\xx$ and $\ww$ \cite{almahairi2018augmented}. Whilst several extensions have been proposed to overcome this problem \cite{zhu2017toward, almahairi2018augmented}, in reality we find that this does not occur in our training setup; we posit that the need to generate realistic radar observations that are capable of tricking the discriminator $\mathcal{G}_\mathrm{x}(\pfx)$ far outweighs $\mathcal{C}_\mathrm{w}(\pfx, \pfw)$, avoiding degeneracy.

\subsection{Learning from Partial Lidar Measurements}
\label{sec:approach-partial}
Another benefit of learning the backward model $\pw(\ww^* | \xx^*)$ is that it allows us to learn from partial measurements $\ww^* \sim p(\ww^*)$ when they are available. This is achieved through an alignment consistency objective,
\begin{gather}
    \mathcal{A}_{\mathrm{w}}(\pfw) = \mathbb{E}_{p(\xx^*, \ww^*)}[\ell(\pfw; \ww^*, \xx^*)]
\end{gather}
where $\ell(\pfw; \ww^*, \xx^*) =\| (\yy - \fw(\ww^*, \zw)) \odot \mathbb{I}(\yy) \|_1$, and $p(\xx^*, \ww^*) = p(\xx^*| \ww^*) p(\ww^*)$ is the joint distribution over real observation-state pairs with $\mathbb{I}(\cdot)$ an element-wise indicator function returning $1$ if the measurement of $\yy_{i, j}$ exists or $0$ otherwise.

Considering a combined training objective,
\begin{equation}
    \label{eq:tran_objective}
    \mathcal{L}(\pfx, \pfw) = \mathcal{G}_\mathrm{x} + \lambda_{\mathrm{gw}} \mathcal{G}_\mathrm{w} +   \lambda_{\mathrm{cx}} \mathcal{C}_\mathrm{x} +  \lambda_{\mathrm{cw}} \mathcal{C}_\mathrm{w} +  \lambda_{\mathrm{aw}} \mathcal{A}_\mathrm{w}
\end{equation}
with hyper-parameters $\bm{\mathrm{\lambda}} =[\lambda_{\mathrm{gw}}, \lambda_{\mathrm{cx}}, \lambda_{\mathrm{cw}}, \lambda_{\mathrm{aw}}]$ used to trade off the relative importance of each term, we are able to train \emph{both} $\px(\xx | \ww)$ and $\pw(\ww^* | \xx^*)$ -- explicitly encouraging $\xx \sim \px(\xx | \ww)$ to generate realistic radar observations in simulation, training on $\ww \in \Ds$, whilst also learning from aligned pairs $\xx, \yy \in \Dr$ when measurements $\yy$ are available.

\section{Experimental Setup}
\label{sec:setup}

\subsection{Self-Supervised Dataset Generation}
\label{sec:data}
In section \ref{sec:formulation} we assumed that we had aligned real-world radar observations and partial state measurements $\Dr= \{(\xx^*, \yy)_n\}_{n=1}^N$ and unaligned but perfectly observed state observations generated in simulation $\Ds=\{\ww_l \}_{l=1}^L$ with which to train our model. We now describe how to attain $(\Dr, \Ds)$ in practice.

\subsubsection{Generating $\Dr$}
\label{sec:setup-Dr}
We generate $\Dr$ from the Oxford Radar RobotCar Dataset \cite{RadarRobotCarDatasetICRA2020, RobotCarDatasetIJRR}. We partition the dataset into train and test sets with the training set being composed of $29$ 10km loops (generating $222420$ observations) with $3$ loops being reserved for testing ($23460$ observations), resulting in an approximate $90:10$ split. Each $\xx^*$ corresponds to the output of a Navtech CTS350x FMCW radar rotating about its vertical axis, down-sampled to a $0.35\text{m}$ resolution and scaled to give $\xx^* \in [-1, 1]^{400 \times 471 }$. This corresponds to a $360^{\circ}$ field-of-view with maximum observable height $5.2\text{m}$. We construct partial height map measurements $\yy$ by combining the output of two HDL32E Velodyne Lidars: as a result of differing sensing frequencies (radar at $4\text{Hz}$ and lidar at $20\text{Hz}$) each $\xx^*$ is matched to multiple lidar scans to maintain accurate labelling. The lidar pointclouds are filtered to coincide with the radar's horizontal field-of-view $(-40^o, 1.8^o)$, before being binned onto a polar grid and labelling each grid cell with the maximum height of any point falling within it. Each $\yy$ is then scaled from the interval $[-2.2\text{m}, 5.2\text{m}]$ to the interval $[-1, 1]$ with any grid cell without a label assigned the value $-1$. Due to the limited number of lasers and range of each lidar, each $\yy$ is only a partial measurement $\yy = m(\ww^*)$ of the true world state, with many cells having no observation attached to them.

\subsubsection{Generating $\Ds$}
\label{sec:setup-Ds}
We generate $\Ds=\{ \ww_l \}_{l=1}^{L}$ utilising the CARLA simulator \cite{dosovitskiy2017carla} by mounting a data collection vehicle with four orthogonal depth cameras at the same height as the radar used to generate $\ww$ ($1.97\text{m}$ above the ground plane), each producing a $\mathbb{R}^{1024\times1024}$ image. These are projected into a dense 3D pointcloud which is then converted to height labels $\ww \in [-1, 1]^{400 \times 471 }$ in a similar approach described in the previous section. In contrast to the the real-world elevation maps generated by lidar we assume that the simulated state observations $\ww$ are dense with each cell potentially observable in radar having a height measurement attached to it. The simulation world was spawned with $200$ vehicles of random types and size and $300$ pedestrians, using town layouts $1$ and $2$. Observations were collected by setting the vehicle into auto-pilot mode and recorded only when the vehicle was moving. The simulation was restarted after $60$ seconds of no movement. Through this approach we generated $10^5$ observations for training. A further $68,400$ were held out for testing.


\subsection{Network Architectures and Training}
Our network architecture and training set-up largely follow that proposed in \cite{CycleGAN2017}. Specifically, we use ResNet generators for $\fx$ and $\fw$ \cite{johnson2016perceptual}, with $2$-strided convolution, $9$ residual blocks, and $2$ up-convolutions before a final $\operatorname{tanh}$ activation. After each convolution, batch normalisation \cite{ioffe2015batch} is applied before a ReLU activation. The variables $\zx$ and $\zw$ are sampled from $\mathcal{N}(\bm{0}, \bm{I})$ before concatenation with $\ww$ and $\xx$ respectively, and passed to $\fx$ and $\fw$ as a 2-channel polar tensor. We utilize patch discriminators for $\dx$ and $\dw$ \cite{ledig2017photo}, sampling generated observations from a pool of $50$ when training as in \cite{CycleGAN2017}. All networks are implemented in PyTorch \cite{NEURIPS2019_9015} and trained for \num{5e5} steps using the Adam optimizer \cite{kingma2014adam} with learning rate \num{2e-4}, $\beta=(0.5, 0.999)$, and a batch size of $1$. In all experiments we set  $\bm{\mathrm{\lambda}} =[1, 10, 10, 10]$ as given in (\ref{eq:tran_objective}). 
\renewcommand{\tabcolsep}{3pt}
\begin{table*}[ht]
\vspace{1em}
\centering
    \begin{tabular}{rcccccccccccc}
    \toprule
    {} &{} & {} & {} & \multicolumn{6}{c}{\textbf{training objective}} & \multicolumn{3}{c}{\textbf{intersection over union}} \\
    {} &  \multicolumn{3}{c}{\textbf{trained on}}   & $\mathcal{A}_\mathsf{x}$ & $\mathcal{A}_\mathsf{w}$ & $\mathcal{G}_\mathsf{x}$ & $\mathcal{G}_\mathsf{w}$ & $\mathcal{C}_\mathsf{x}$ & $\mathcal{C}_\mathsf{w}$ &  free &   occ &  mean \\
    \midrule
    \textbf{benchmark} & & &  & & & & & & & & \\
    real world &  - &    - & - & - & - & - & - & - & - & 0.856 & 0.553 & 0.705 \\
    \textbf{ours} &  & & & & & & & & & &\\
    (a)  & $\xx^*$ & $\yy$ & - & \tick &  - & - & - & - & - & 0.396 (0.00) & 0.275 (0.00) & 0.335 (0.00) \\
    (b)  & $\xx^*$ & $\yy$ & - & \tick &  - & \tick & - & - & - & 0.558 (0.11) & 0.221 (0.03) & 0.389 (0.07) \\
    (c)  & $\xx^*$ & - &$\ww$ &  - &  - & \tick & - & - & - & 0.385 (0.07) & 0.148 (0.01) & 0.266 (0.04) \\
    (d)  & $\xx^*$ & - & $\ww$ &  - & - & \tick & \tick & \tick & \tick & 0.845 (0.02) & 0.262 (0.02) & 0.553 (0.02) \\
    (e) &  $\xx^*$ & $\yy$ & $\ww$ & - & \tick & \tick & \tick & \tick & \tick & \textbf{0.872} (0.01) & \textbf{0.455} (0.01) & \textbf{0.664} (0.01) \\
    \bottomrule
    \vspace{-1.5em}
    \end{tabular}
\caption{Radar simulation performance: real-world mIoU performance for segmentation models trained by different radar sensor models in simulation as outlined in Section \ref{sec:approach} and discussed in Section \ref{sec:result-radar-sim} (averaged over four random seeds presented with standard deviations)} \label{tab:iou}
\end{table*}

\begin{figure*}
    \centering
    \includegraphics[width=\linewidth]{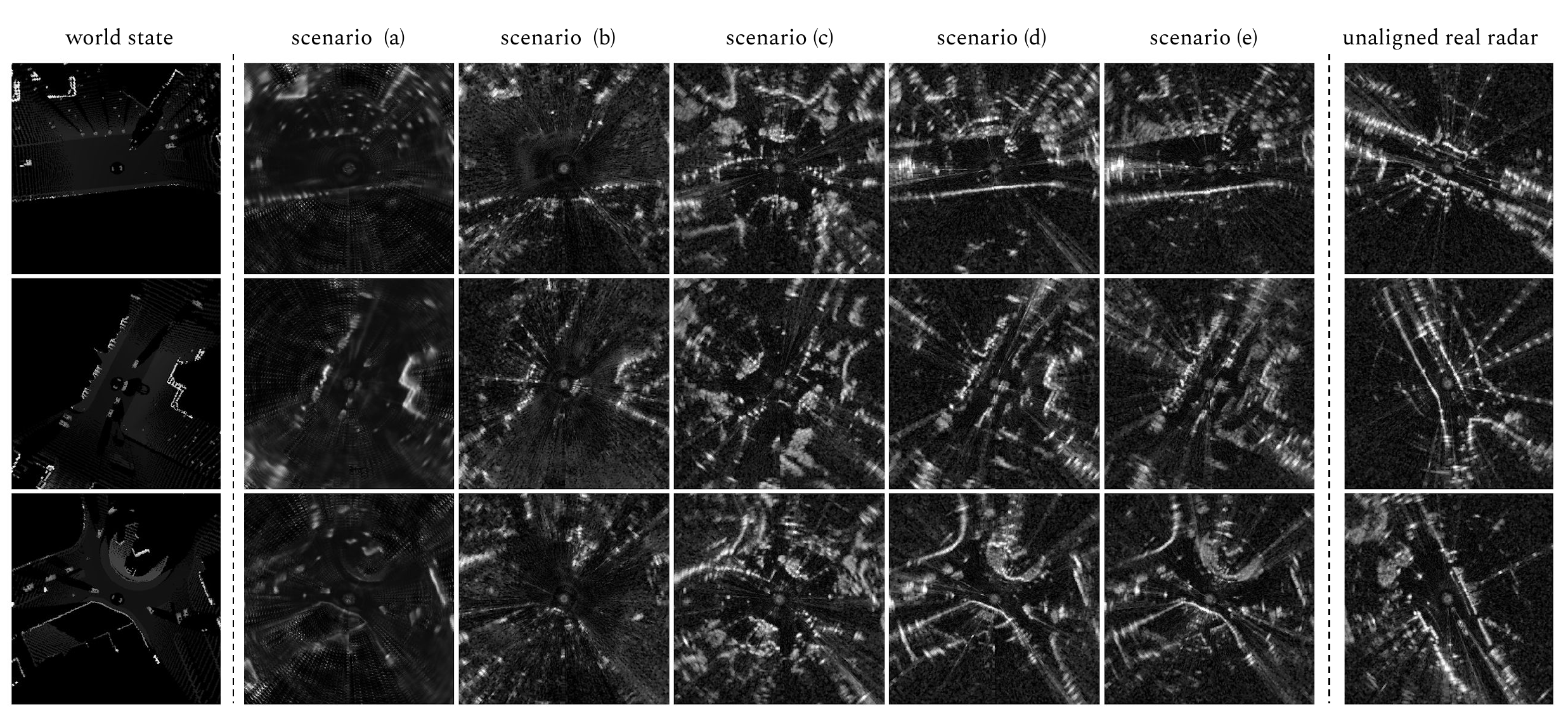}
    \caption{Simulated radar for training scenarios given in Table \ref{tab:iou} generated from elevation maps given in the first column. \emph{Unaligned} real radar observations $\xx^*$ are also shown for reference (last column). Simulators trained using only real data, as in (a) and (b), fail to synthesise realistic radar in simulation. Whilst at first glance simulators trained using only an adversarial criterion between simulated and real radar appear realistic, on closer inspection they are poorly aligned to the world state as can be seen in (c). Enforcing cyclical consistency helps to remedy this as can be seen in (d). The most realistic radar observations correspond to simulators trained with the full training objective, as in (e), and backed up by Table \ref{tab:iou}.}
    \label{fig:setups}
    \vspace{-1.7em}
\end{figure*}

\subsection{Evaluation}
\subsubsection{Radar Simulation}
\label{sec:approach-eval}
One of the central motivations for our approach was to develop a radar sensor model $\px(\xx | \ww)$ to train new models $\fo(\xx)$ in simulation that generalise to the real world. With this in mind, to assess the realism of radar observations $\xx \sim \px(\xx | \ww)$, we train a model $\fo(\xx)$ in simulation minimising $\mathbb{E}_{\px(\xx, \oo)}[\ell(\oo, \fo(\xx))]$ with respect to $\alpha$ where $\ell(\cdot, \cdot)$ is a loss between the actual and predicted target. We then assess the realism of $\px(\xx | \ww)$ as, $\mathbb{E}_{p(\xx^*, \oo^*)}\left[m(\oo^*, \fo(\xx^*|\pfx))\right]$ evaluating the performance of the trained model $\fo(\xx^*|\pfx)$ in the real world with metric $m(\cdot, \cdot)$. Specifically, we consider training a model to segment the world into occupied, free, and unknown space. To successfully train $\fo(\xx)$ this task requires $\xx \sim \px(\xx | \ww)$ to replicate realistic noise artefacts (which $\fo(\xx)$ must learn to overcome) whilst ensuring the mapping from world state $\ww$ to radar observation $\xx$ is as faithful as possible to the real-world mapping $\ww^*$ to $\xx^*$. 

We train and evaluate the segmentation model using datasets $\Ds'=\{(\xx, \oo)\}_{n=1}^N$ and $\Dr'=\{(\xx^*, \oo^*)\}_{n=1}^N$, generated from the sets held out in sections \ref{sec:setup-Dr} and \ref{sec:setup-Ds}. In both cases labels are generated automatically from either partial $\yy$ or full $\ww$ measurements of the world state in a similar approach to \cite{weston2019probably}: after extracting the ground plane, any cell with a height measurement is labelled as occupied, all cells before the first return are labelled as free, and anything else is labelled unknown. For all segmentation networks $\fo$, a U-Net architecture \cite{ronneberger2015u} is used with $6$ levels, doubling the features and halving spatial resolution at each level, starting with $8$ features at the input and allowing information to flow from encoder to decoder using skip connections. Each model is trained with a batch size of $8$ using the Adam \cite{kingma2014adam} optimizer (learning rate \num{1e-3}) to minimise the cross-entropy loss, with an additional weighting of $50$ applied to the occupied class to account for class imbalance. The model attaining the highest IoU over $4$ epochs of training (tested on a $10\%$ hold out dataset after each epoch) is used for evaluation -- usually the first or second.

In keeping with \cite{geiger2012we} and as proposed in the Pascal Voc Challenge \cite{hoiem2009pascal} each model is evaluated using the mean Intersection over Union metric \cite{everingham2010pascal} (mIoU)
\begin{gather}
    \label{eq:mIoU}
    \mathcal{M}(\pfx) = \frac{1}{2} \sum_{c} \left[ \frac{\mathrm{TP}(c)}{\mathrm{FP}(c) + \mathrm{FN}(c) + \mathrm{TP}(c)} \right]
\end{gather}
where the true positives $\mathrm{TP}$, false positives $\mathrm{FP}$ and false negatives $\mathrm{FN}$ are determined for each class (occupied and free) comparing $\oo^*_{i, j}$ and $\fo(\xx)_{i, j}$ at each index $(i, j)$ across the entire dataset $\Dr'=\{(\xx^*, \oo^*)\}_{n=1}^N$. Any cell that is predicted unknown but labelled as free or occupied is counted as a false negative.

\subsubsection{Evaluating the Backward Model}
\label{sec:setup-height-eval}
The heights predicted by our backward model $\ww^* \sim \pw(\ww^* | \xx^*)$ are evaluated using absolute mean error between $\ww^* \sim \pw(\ww^* | \xx^*)$ and partial state measurements $\yy$ over the test set $\Dr'$ from section \ref{sec:setup-Dr}, evaluated only where measurement $\yy_{i, j}$ exist. As a result of the class imbalance between height labels corresponding to the ground plane and targets in the scene, we evaluate this metric over occupied and free space independently, presenting both alongside their average. All height evaluations were run for four models trained with different random seeds, from which we computed averages and error bounds presented as standard deviations.

\section{Results}
\label{sec:results}
\subsection{Radar Simulation}
\label{sec:result-radar-sim}
To assess the realism of radar observations simulated through our approach $\xx \sim \px(\xx | \ww)$, we consider how models trained using $\px(\xx | \ww)$ perform in the real world using the evaluation method proposed in section \ref{sec:approach-eval}. The results for an ablation of possible training setups is given in Table \ref{tab:iou} with a qualitative comparison given in Figure \ref{fig:setups}. 
In (a) and (b) we assess whether it is possible to learn to simulate realistic radar observations given a simulated elevation map $\ww$ whilst only training on real-world measurements $(\xx^*, \yy) \in \Dr$ as proposed in section \ref{sec:approach-real_only}. In (a) we train $\px(\xx | \ww)$ to regress to $\xx^*$ directly from partial state measurements minimising (\ref{eq:tran_objective}) whilst in (b) we add an additional adversarial loss $\mathcal{G}_{\mathrm{x}}$ introducing a discriminator $\dx(\xx)$ to distinguish between real and simulated radar observations (similar to \cite{pix2pix2017}). Both (a) and (b) lead to models that poorly generalise to the real world corresponding to a mIoU of $0.335$ and $0.389$ respectively, and as seen qualitatively in Figure \ref{fig:setups}.

In (c) and (d) we consider learning using only unaligned observations $(\xx^*, \ww)$ assuming that partial state measurements $\yy$ are unavailable. Specifically, in (c) we train a model trained using only an adversarial loss $\mathcal{G}_\mathrm{x}$ between real $\xx^*$ and simulated radar observations $\xx \sim \ww^*$ as proposed in section \ref{sec:approach-adversarial} whilst in (d) we add in cyclical consistency constraints $\mathcal{C}_\mathrm{x}$ and $\mathcal{C}_\mathrm{w}$, also learning the backward model $\pw(\ww^*|\xx^*)$ as proposed in section \ref{sec:approach-cyc}. We find that (c) leads to models that perform poorly in the real world in this instance achieving a mIoU of only $0.266$; training only on adversarial criteria leads to simulated radar observations that are poorly aligned to the real-world state as can be seen in Figure \ref{fig:setups}. Additionally modelling the backward model and imposing cyclical consistency, as in (d), allows us to encourage alignment between the world state and simulated radar observation -- boosting performance to $0.553$ and significantly outperforming (a) and (b).

However, our best performing model (e) is trained using the full training objective given in (\ref{eq:tran_objective}) as proposed in section \ref{sec:approach-partial}. In addition to producing the most realistic simulated observations $\xx$ in Figure \ref{fig:setups}, in this case we are able to train a model in simulation achieving a mIoU of $0.664$, only $4$ percentage points off the same model trained directly on real data in the same domain as the test set $0.705$.

\subsection{Height Inference}
\label{sec:eval_height_inference}

In addition to improving the realism of the radar sensor model, as demonstrated in the last section, the backward model $\pw(\ww^*| \xx^*)$ learnt as part of the same training setup can be used to infer the underlying elevation state of the world given a real world radar observation $\xx^*$.

\begin{figure}[ht]
    \centering
    \includegraphics[width=0.9\linewidth]{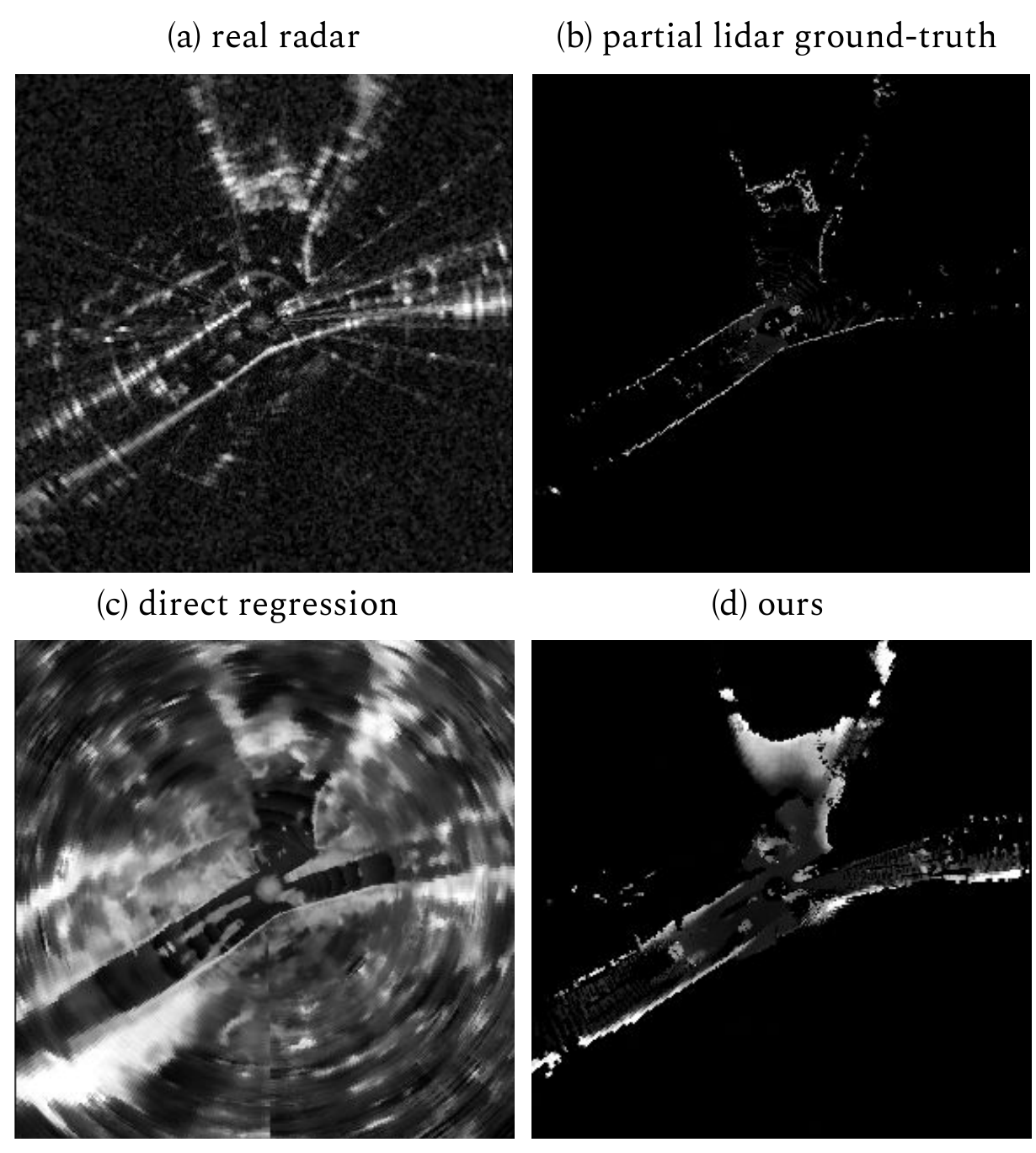}
    \caption{Predicted elevation maps from real-world radar. As can be seen in (c) models learnt by directly regressing to partial height maps $\yy$ poorly generalise to regions without labels. Adding in an additional adversarial loss forces the predicted elevation to look more like simulated height maps allowing the model in (d) to generalise to regions of space for which labels do not exist. }
    \label{fig:height_inference}
    \vspace{-1.7em}
\end{figure}

We evaluate the quality of the heights predicted by $\pw(\ww^*| \xx^*)$ using the evaluation procedure described in section \ref{sec:setup-height-eval}. We use the MAE error as compared to the partial height measurements $\yy$ made in lidar evaluated separately for both both free and occupied space. The results of this process are presented in Table \ref{tab:height_results} for several different training configurations. In (a) we consider just training $\pw(\ww^* | \xx^*)$ by enforcing only cyclical consistency as proposed in section \ref{sec:approach-cyc} whilst in (b) we consider training $\pw(\ww^* | \xx^*)$ using the full training criteria given in equation (\ref{eq:tran_objective}). By imposing alignment in (b) the system is able to more accurately infer the elevation map $\ww^*$ with MAE $23\text{cm}$ -- compared to a $39\text{cm}$ accuracy for (a). Whilst (b) performs slightly worse than our benchmark of $13\text{cm}$ we find that as a result of the additional adversarial constraint $\mathcal{G}_{\textrm{w}}$, we are able to generalise to regions outside of the range of lidar (unlike our benchmark) as can be seen in Figure \ref{fig:height_inference}.
\renewcommand{\tabcolsep}{3pt}
\begin{table}[ht]
    \centering
  \begin{tabular}{lccccccc}
\toprule
{} &  & & & & \multicolumn{3}{c}{\textbf{Mean Absolute Error (cm)}} \\
{} & data & $\mathcal{A}_\mathsf{z}$ & $\mathcal{G}_\mathsf{w}$ & $\mathcal{C}_\mathsf{w}$ & free & occ &  mean \\
\midrule
\textbf{Benchmark} & & & & & & \\
direct regression  &  $\Dr$ &  \tick & - & - & 18.7 & 7.5 & 13.1 \\
\textbf{Ours} & & & & & & \\
(a)         &  $\Dr,\Ds$ &  - &  \tick &  \tick & 3.9 (0.4) &  74.2 (1.8) &  39.0 (0.9) \\
(b) &  $\Dr,\Ds$ &   \tick & \tick & \tick &    4.1 (0.6) &  41.1 (5.7) &  22.6 (2.6) \\
\bottomrule
\end{tabular}
\vspace{0.5em}
\caption{MAE for the heights predicted by the backward model evaluated against partial height measurements generated in lidar as proposed in Section \ref{sec:eval_height_inference}. Our benchmark corresponds to a model trained to regress directly to lidar measurements. (Averaged over four random seeds and presented with standard deviations.)} 
\label{tab:height_results}
\vspace{-2em}
\end{table}

\section{Conclusion}
This work demonstrates that it is possible to develop a data-driven approach to radar-simulation capable of training downstream systems that are readily deployable in the real world. By simulating radar sensor observations from elevation maps we are able to interface with existing simulators already capable of synthesising complex real-world scenes. We adopt an inherently stochastic and data-driven approach, capturing a mapping from state to radar sensor model (alongside sensor noise). We learn our approach from real radar measurements, simulated elevation maps, and partial elevation measurements generated in lidar. To encourage our model to simulate realistic radar observations, we adopt an adversarial approach model the backward mapping 
to further constrain learning through cyclical consistency losses and partial alignment to real-world elevation maps. 

Using our approach to train a segmentation system in simulation, we find that when deployed in the real world, the system is able to operate with a mIoU of $0.664$ performing comparably to a model trained in the real world only. To the best of our knowledge this is the first time that the feasibility of training models in simulation has been demonstrated in radar. The backward model learnt as part of the same training setup can be used to infer the height state of the world with an accuracy of $23\text{cm}$, using only partial elevation measurements, whilst generalising to regions of space for which no labels exist. 

Whilst our model is able to successfully train segmentation models in simulation that partition the world into occupied, free, and unknown space, early experiments (on a limited test set) found that partitioning occupied space into finer grained classes was significantly more challenging. This constitutes an interesting area for future research.

\newpage
\clearpage

\bibliography{bibliography}
\bibliographystyle{ieeetr}

\end{document}